\documentclass[11pt]{article}
\pdfoutput=1
\usepackage[margin=1.4in]{geometry}
\usepackage{amsmath,amsthm,enumerate,graphicx,amsfonts}
\usepackage[utf8]{inputenc}
\usepackage{color}

\theoremstyle{plain}

\usepackage{amsmath,amsfonts,bm}

\def\eqref#1{equation~\ref{#1}}

\def\1{\bm{1}}

\def\vb{{\bm{b}}}

\def\vz{{\bm{z}}}

\def\mA{{\bm{A}}}
\def\mB{{\bm{B}}}

\def\mD{{\bm{D}}}
\def\mE{{\bm{E}}}

\def\mH{{\bm{H}}}

\def\mK{{\bm{K}}}

\def\mM{{\bm{M}}}

\def\mO{{\bm{O}}}
\def\mP{{\bm{P}}}
\def\mQ{{\bm{Q}}}

\def\mU{{\bm{U}}}
\def\mV{{\bm{V}}}
\def\mW{{\bm{W}}}
\def\mX{{\bm{X}}}

\DeclareMathAlphabet{\mathsfit}{\encodingdefault}{\sfdefault}{m}{sl}
\SetMathAlphabet{\mathsfit}{bold}{\encodingdefault}{\sfdefault}{bx}{n}

\usepackage{hyperref}
\usepackage{url}
\usepackage{algorithm}
\usepackage{algpseudocode}
\usepackage{xspace}
\usepackage{booktabs}

\title{GUIDE: Guided Initialization and Distillation of Embeddings}

\author{Khoa Trinh, Gaurav Menghani \& Erik Vee  \\
Google Research \\
Mountain View, CA \\
\texttt{\{khoatrinh,gmenghani,erikvee\}@google.com} 
}

\newcommand{\guide}{\textsc{GUIDE}\xspace}

\begin{document}

\maketitle

\begin{abstract}
Algorithmic efficiency techniques such as distillation (\cite{hinton2015distillation}) are useful in improving model quality without increasing serving costs, provided a larger teacher model is available for a smaller student model to learn from during training. Standard distillation methods are limited to only forcing the student to match the teacher's outputs. Given the costs associated with training a large model, we believe we should be extracting more useful information from a teacher model than by just making the student match the teacher's outputs.

In this paper, we introduce \guide (Guided Initialization and Distillation of Embeddings). \guide can be considered a distillation technique that forces the student to match the teacher in the parameter space. Using \guide we show 25-26\% reduction in the teacher-student quality gap when using large student models (400M - 1B parameters) trained on $\approx$ 20B tokens. We also present a thorough analysis demonstrating that \guide can be combined with knowledge distillation with near additive improvements. Furthermore, we show that applying \guide alone leads to substantially better model quality than applying knowledge distillation by itself.

Most importantly, \guide introduces no training or inference overhead and hence any model quality gains from our method are virtually free. 
\end{abstract}

\section{Introduction}
With the advent of Large Language Models (LLMs), a naive way to scale model performance is via increasing the number of model parameters (increasing width and/or depth), and/or training on more tokens. Better model quality achieved through such methods comes at the cost of increased model footprint metrics such as inference latency, memory, etc., and training costs.

Such a naively scaled up model might perform strongly on benchmarks, but could also have prohibitively large latency and/or memory requirements, which might make it infeasible to be trained or deployed at large scale. Therefore, it is of crucial to pay attention to \emph{model efficiency} at the same time as model quality. Many recent works cover various aspects of model efficiency such as algorithmic efficiency techniques (\cite{menghani2023efficientdl}), hardware support (\cite{sze2017survey}), and best practices (\cite{dehghani2022efficiency}).

For the purpose of this work, we are specifically interested in algorithmic efficiency techniques like Distillation \cite{hinton2015distillation}, which helps a smaller \emph{student} model learn from a larger \emph{teacher} model during training, without having any impact on the model's footprint metrics (parameters, latency, memory, etc.). Since its introduction, distillation has found widespread use across models of various kinds (\cite{kim2023embeddistill, sanh2019distilbert}).

In this paper, we introduce \emph{Guided Initialization and Distillation of Embeddings}, or \guide, which is a novel approach to improve model quality without any impact on the model's footprint metrics. We know that standard knowledge distillation leads to progressive transfer of knowledge from the teacher to the student, by forcing the student to match the teacher's predictions. On the other hand, \guide is designed so that the student also utilizes the teacher model's parameters to have a strong initialization \emph{guided} by the teacher model. We can view \guide as distillation in the parameter space, without the need for the teacher model to label a dataset.

\section{Related Work}{\label{sec:related_work}}
\guide is an algorithmic efficiency technique, as mentioned earlier, closely related to and combinable with distillation \cite{hinton2015distillation} and its variants \cite{sanh2019distilbert}. However, unlike distillation, \guide does not change the loss function and neither does it need the teacher model to label its training dataset.

A closely related work \cite{xu2023init} pitches a very similar idea on very small models (5M parameter student and 22M parameter teacher models). However, we show that their method does not scale to large models of size up to 1B parameters.

Transferring knowledge in the parametric space from the teacher to the student has been attempted in \cite{sanh2019distilbert, sum_dist} with positive results in LLMs. Unfortunately, in these works, the authors assume that both the teacher and the student share the same model's embedding dimension, which allows them to let the student copy whole chosen layers from the teacher. This requirement is too restrictive because very large teacher models often have a much larger embedding dimension than the student.

In \cite{zhong2024seeking}, the authors introduce a sensitivity-based technique to extract and align parameters between teacher and student LLMs. However, their approach requires adding another LoRA module and training that module to select the teacher's relevant weights. Similarly, in \cite{xia2024sheared}, the proposed pruning method tries to solve a constrained optimization problem, requiring us to learn pruning masks in order to decide which parameters should be kept from the teacher model. \cite{weight_dist} propose that a ``Parameter Generator'' can be trained to first predict the student's weights. Then we can continue finetuning the student model. However, this incurs an inference cost from the teacher while training the generator. Thus, all of these methods are quite expensive in practice. On the other hand, GUIDE does not require any new parameters or additional training.

There have also been attempts to use the weights of smaller students to initialize and train a larger teacher model \cite{chen-etal-2022-bert2bert, Samragh2024ScalingSA}. In this work, we only transfer knowledge from a trained teacher to a smaller student.

\cite{frankle2019lottery} shows that well-trained large models may contain \emph{subnetworks} that can perform almost as well as the original larger models. They provide a recipe that can be followed to extract smaller subnetworks from simple fully connected networks and convolutional neural networks. 

They also demonstrate that when trained from scratch, the subnetworks do not attain the same performance as when extracted from larger models. This aligns with our results, where we empirically confirm that models using \guide to parametrically distill from a larger teacher perform significantly better than initializing from scratch.

Another method \cite{wortsman2022soup} proposes averaging the weights of multiple models with the same architecture, when fine-tuned on the same dataset and initializing from the same pre-training checkpoint (referred to as `Model Soup'). This is akin to ensembling in the parameteric space.  

\cite{devvrit2024matformer} proposes a method that can be used to train nested smaller models within a larger model. This can be useful when a new larger model is being trained froms scratch, but is applicable for leveraging an existing large model that was trained without this technique.

Other common efficiency techniques such as quantization \cite{krishnamoorthi2018quantization, jacob2018quantization}, sparsity \cite{gale2019sparsity}, architectural techniques \cite{menghani2025laurel}, etc. are orthogonal to our work can be applied in conjunction with GUIDE without any adverse effects.



\section{Guided Initialization and Distillation of Embeddings (GUIDE)}
In this section, we introduce GUIDE, a simple and fast approach which directly transfers knowledge from a teacher model to the student model by using the pretrained teacher to initialize the student model's parameters. 

\subsection{Preliminaries}
In the standard distillation setting, we have a teacher model $T$ and we want to distill $T$ into a smaller student model $S$. Generally speaking, the teacher has more parameters and layers and performs better than the student. Here we assume that $S$ and $T$ are traditional transformers to be trained on the same dataset with the same context length. For ease of notation, we shall use $S$ and $T$ as subscripts of the relevant variables, tables, and weight matrices to indicate whether they are from the student or teacher model. Let $d, n, h, \ell$, and $f$ be the model dimension, the number of transformer blocks, the number of attention heads, the key/query/value dimension, and the MLP inner dimension, respectively. Recall that $d_S = h_S \ell_S$ and $d_T = h_T \ell_T$. In this work, we also assume that $d_S \leq d_T, h_S \leq h_T, $ and $\ell_S \leq \ell_T$.

Next, let us introduce some notation needed to describe our method and briefly recap how transformer-like models work. Note that we only describe the relevant details here while skipping others (e.g. layer normalization, causal mask, etc.). The reader is referred to \cite{transformer} for a full description of the transformer architecture. Let $\mE$ and $\mP$ be the embedding table and the positional encoding table of size $m \times d$ where $m$ is the vocabulary size. Given an input sequence of token indices $\vz$ of length $L$ (i.e., context length), the input embedding $\mX \in \mathbb{R}^{L \times d}$ is computed by looking up tokens from $\mE$ and adding the positional encodings from $\mP$:
$$ \mX_i := \mE_{\vz_i} + \mP_i.  $$
Then $\mX$ is passed through a sequence of $n$ multi-head self-attention blocks where the output of one block is added to the original input and becomes the input of the next block. Each block has $h$ attention heads and an MLP layer that work as follows. Slightly abusing notation, for each attention head $i \in \{1, \ldots, h\}$, define learnable projection matrices $\mW_i^Q, \mW_i^K, \mW_i^V \in \mathbb{R}^{d \times \ell}$. We then project $\mX$ onto these to get the query, key, and value matrices: $\mQ_i = \mX \mW_i^Q, \mK_i = \mX \mW_i^K, \mV_i = \mX \mW_i^V$. The attention scores is computed as
$$ \mA_i := \text{softmax}\left(\frac{\mQ_i \mK_i^T}{\sqrt{\ell}}   \right) \in \mathbb{R}^{L \times L}.$$
The scores are then used as weights to compute the contextual embeddings:
$$\mH_i := \mA_i \mV_i \in \mathbb{R}^{L \times \ell}.$$
The head outputs are then concatenated and projected onto a learnable output projection matrix $\mW^O \in \mathbb{R}^{d \times d}$:
$$\mO := \text{Concat}(\mH_1, \ldots, \mH_h)\mW^O \in \mathbb{R}^{L \times d}.$$
Finally, $\mO$ is pushed through an MLP layer giving the block output:
$$\mB := \phi(\mO \mW^{(1)} + \vb^{(1)})\mW^{(2)} + \vb^{(2)} \in \mathbb{R}^{L \times d}, $$
where $\mW^{(1)} \in \mathbb{R}^{d \times f}, \mW^{(2)} \in \mathbb{R}^{f \times d} $, $\vb^{(1)} \in \mathbb{R}^{f}$, and $\vb^{(2)} \in \mathbb{R}^{d}$ are learnable, $\phi$ is an activation function (e.g., GELU), and the $+$ operator is applied in row-wise manner.

\subsection{Initial Attempts}

In this study, we explore new ways to improve standard knowledge distillation. Can we extract even more useful information from the teacher to help train a better student? A popular approach, in the same spirit as the original distillation work of \cite{romero2014fitnets}, would be to exploit the teacher's final token representations (that is, the output of the last block used to compute the logits) $\mB_{T, h_T}$. In fact, matching the student's and teacher's token embeddings has yielded positive results in \cite{tinybert, kim2023embeddistill}. Here we encounter the first challenge: the dimensional mismatch between the teacher and student. A common approach to this problem is to introduce another learnable projection matrix $\mM \in \mathbb{R}^{d_T \times d_S}$ and optimize the following additional MSE loss:
$$ \text{loss}_{\text{embedding}} := \text{MSE}(\mB_{S, n_S}, \mB_{T, n_T} \times \mM), $$
where the parameters in $\mB_{T, n_T}$ are frozen \cite{romero2014fitnets}. Unfortunately, this approach not only incurs the teacher's inference cost in the training proccess, but also does not seem to work consistently when pretraining decoder-only LLMs. In our experiments, adding this embedding loss actually hurts the student's performance.

We want to directly transfer knowledge from the teacher to the student via an ``almost-free'' weight initialization method. Basically, the weights of the pretrained teacher can be used as a good starting point for the student. As mentioned in Section \ref{sec:related_work}, notable works on this subject include \cite{xu2023init, sum_dist, sanh2019distilbert}. The weight initialization methods consist of two main components: layer selection and weight selection. The former is a strategy to select teacher's layers to be matched to student's layers. For a given pair of teacher's and student's layers, the latter helps select a subset of teacher's weights to be used as initial weights for the student's layer.

\subsubsection{Uniform Selection by \cite{xu2023init}}
For isotropic architectures (e.g., transformer models), \cite{xu2023init} suggests the so-called ``first-$N$ selection`` for layer selection. Basically, the first $N$ layers of the teacher will be used as the initialization source. The authors also mention taking evenly-spaced layers in teacher as an alternative. Note that this approach has also been tried in \cite{sum_dist}. For weight selection, they propose the ``Uniform Selection'' strategy as follows. Suppose that we want to initialize the student's weight tensor $\mW_S$ with shape $s_1 \times s_2 \times \ldots s_n$ from $\mW_T$ with shape $t_1 \times t_2 \times \ldots \times t_n$ where the indices are 0-indexed.

\begin{algorithm}[h]
\caption{\textsc{GetEvenlySpacedIndices}($m$, $n$) \textit{~~~~ Note: $m,n$ are integers and $n \geq m > 1$}} \label{alg:getEvenIndices}
\begin{algorithmic}[1]
\State $I \gets \left(0, \frac{m-1}{n-1}, \frac{2(m-1)}{n-1}, \ldots, \frac{(m-1)^2}{n-1}\right)$
\State Round numbers in $I$ to the closest integer (half-integers are rounded down)
\State Return $I$
\end{algorithmic}
\end{algorithm}

To construct $\mW_S$, for each dimension $i \in [1,n]$, we select indices returned by \textsc{GetEvenlySpacedIndices}($s_i$, $t_i$) along the $i$-th dimension of $\mW_T$. While this procedure is simple and can be applied to different model architectures, it is agnostic to how transformers actually work. 

To the best of our knowledge, none of the current works take advantage of the specific structure of transformers in weight initialization. Here we shall try to extract even more information from the teacher's embedding table and the first transformer layer, which directly takes the token embeddings from the table as input.

\subsubsection{Low-rank approximation to preserve pairwise dot-products}
Let us start with the embedding table of the teacher $\mE_T$. One natural idea would be to replace the student's embedding table by $\mE_T \mM$ where $\mE_T$ is frozen and $\mM \in \mathbb{R}^{d_T \times d_S}$ is learnable as above. The drawback of this approach is the extra parameters in the projection $\mM$. Moreover, learning this projection does not seem to yield significant improvements in practice. Thus, we want to directly initialize $\mE_S$ by using $\mE_T$. One idea would be to find $\mE_S$ such that the pairwise dot products between tokens in the dictionary are preserved:
$$\mE_S \mE_S^T \approx \mE_T \mE_T^T,$$ 
which is a low-rank approximation problem. Let $\mD := \mE_T \mE_T^T$. Observe that $\mD$ is symmetric positive semidefinite. So we can find the spectral eigendecomposition of $\mD$:
$$ \mD = \mU \mathbf{\Lambda} \mU^T, $$
where $\mU$ is orthogonal and $\mathbf{\Lambda} = \text{diag}(\lambda_1, \ldots, \lambda_L)$ with $\lambda_1 \geq \lambda_2 \geq \ldots \geq \lambda_L \geq 0$. Let $\mU_{d_S} := \mU_{:, :d_S}$ and $\mathbf{\Lambda}_{d_S} := \text{diag}(\lambda_1, \ldots, \lambda_{d_S})$. Then according to the Eckart–Young–Mirsky theorem, taking
$$ \mE_S := \mU_{d_S} \mathbf{\Lambda}_{d_S}^{1/2}$$
gives the best solution among all matrices of rank at most $d_S$ to minimize $\|\mE_S \mE_S^T - \mD\|_\text{F}$.

Note that the above low-rank approximation method is a parametric knowledge transfer that does not require the teacher to participate in the student training. However, it is not trivial to apply the same approach to the transformer blocks and/or to make these blocks work in sync with the embedding table constructed this way. For example, suppose that we want to extract knowledge from some attention head $i$ of a teacher's block. One straightforward idea is to let the student learn the teacher's attention scores $\mA_{T,i}$. However, these scores depend not only on parametric projections $\mW_{T,i}^Q$ and $\mW_{T,i}^K$, but also on the teacher's input embeddings $\mX$. Traditionally, this would require the use of an additional loss, for example, $\|\mA_{S,i}-\mA_{T,i}\|_2$. Then we still have to deal with $\mW_{T,i}^V$, $\mW_{T,i}^O$, and the teacher's MLP layer.

\subsection{Our Approach}

We now propose a simple and efficient method for the weight initialization of transformer models. The main idea here is to take $\mE_S$ as the PCA compression of $\mE_T$. By projecting $\mE_T$ onto its PCA projection matrix $\mM \in \mathbb{R}^{d_T \times d_S}$, we retain the most important patterns in $d_S$ dimensions. Unlike previous pruning approaches that ``approximate'' the teacher's embedding table by picking some subset of the features, here we aim to extract as much information from it as possible. 

\begin{figure}[h]
\centering 
\vspace{.1in}
\includegraphics[width=0.6\textwidth]{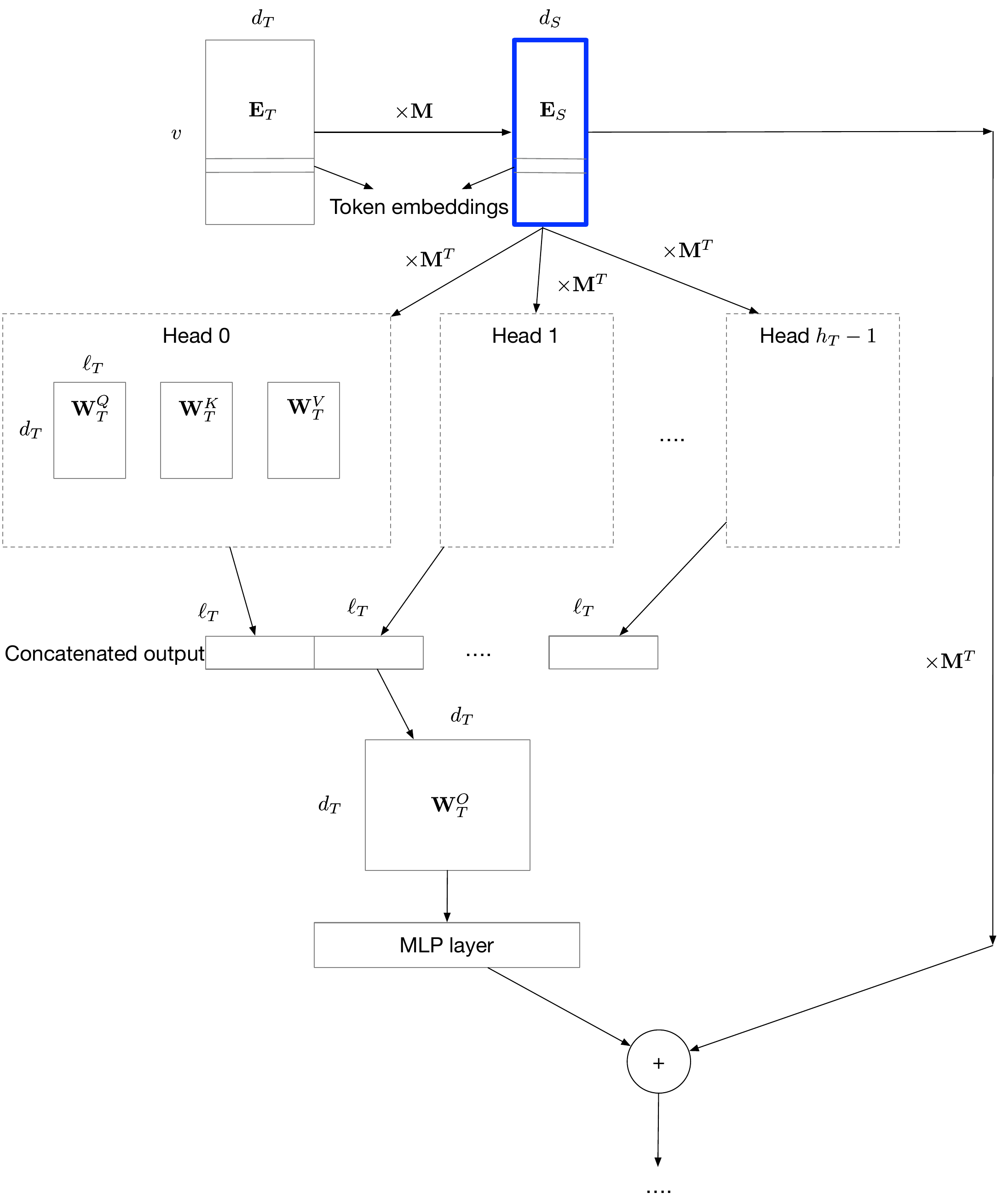}
\vspace{.1in}
\caption{
The teacher transformer can be ``approximated'' by using a PCA-compressed embedding table along with reconstructed embeddings.
}
\label{fig:algo0}
\end{figure}

More importantly, $\mM$ can be used as a ``bridge'' connecting the student embedding space and the teacher embedding space. Note that multiplying any student's embedding by $\mM^T$ would reconstruct the original teacher's embedding. Therefore, by projecting the input sequence $\mX$ onto $\mM^T$ before pushing it into the first block, the teacher model now works harmonically with the new smaller embedding table $\mE_S$ as illustrated in Figure \ref{fig:algo0}. The query, key and value matrices in this case are $\mQ_0 = (\mX\mM^T)\mW_0^Q, \mK_0 = (\mX\mM^T)\mW_0^K,$ and $\mV_0 = (\mX\mM^T)\mW_0^V$. By the associativity of matrix multiplication, we can also absorb $\mM^T$ into $\mW_0^Q,$ and $\mW_0^K, \mW_0^V$, transforming their shape to $d_S \times \ell_T$. For the rest of the block, we can now use Uniform Selection to reduce the remaining dimensions of $\ell_T, d_T,$ and $f_T$ to $\ell_S, d_S,$ and $f_S$, respectively. 

The details of GUIDE are shown in Algorithm \ref{alg:pca_guide}.

\begin{algorithm}[h]
\caption{\textsc{GUIDE}($S$, $T$) \textit{~~~~\# {$S,T$ represent the student and teacher models}} \label{alg:pca_guide}}
\begin{algorithmic}[1]
\State $\mE \gets \text{Concat}(\mE_S, \mP_S)$ \textit{~\# concatenate in row-wise manner}
\State $\mU, \mathbf{\Sigma}, \mV \gets \textsc{SVD}(\mE)$
\State $\mM \gets \mV[:, :d_S]$  \textit{~~~\# the PCA projection matrix, taking the most $d_S$ important directions}
\State Initialize $\mE_S \gets \mE_T \mM$ and $\mP_S \gets \mP_T \mM$
\State Initialize $Q,K,V$ projection matrices of the first block:
$\mW_{S,0}^Z \gets \mM^T\mW_{T,0}^Z$ for each $Z \in \{Q,K,V\}$
\State Let $D, D_h, H,$ and $ F$ be the outputs of \textsc{GetEvenlySpacedIndices} when applying to pairs $(d_S, d_T), (\ell_S, \ell_T), (h_S, h_T),$ and $ (f_S, f_T)$ respectively.
\State $\mW_{S,0}^Z \gets \mW_{S,0}^Z[:, D_h]$ for each $Z \in \{Q,K,V\}$
\State Reshape $\mW_{T,0}^O$ to $h_T \times \ell_T \times d_T$ 
\State Initialize $\mW_S^O \gets \mW_{T,0}^O[H, D_h, D]$ 
\State Reshape $\mW_{S,0}^O$ back to $d_S \times d_S$
\State $\mW_{S,0}^{(1)} \gets \mW_{T,0}^{(1)}[D,F], \vb_{S,0}^{(1)} \gets \vb_{T,0}^{(1)}[F]  $
\State $\mW_{S,0}^{(2)} \gets \mW_{T,0}^{(2)}[F,D], \vb_{S,0}^{(2)} \gets \vb_{T,0}^{(2)}[D]$
\end{algorithmic}
\end{algorithm}

\begin{figure}[h]
\centering 
\vspace{.1in}
\includegraphics[width=0.6\textwidth]{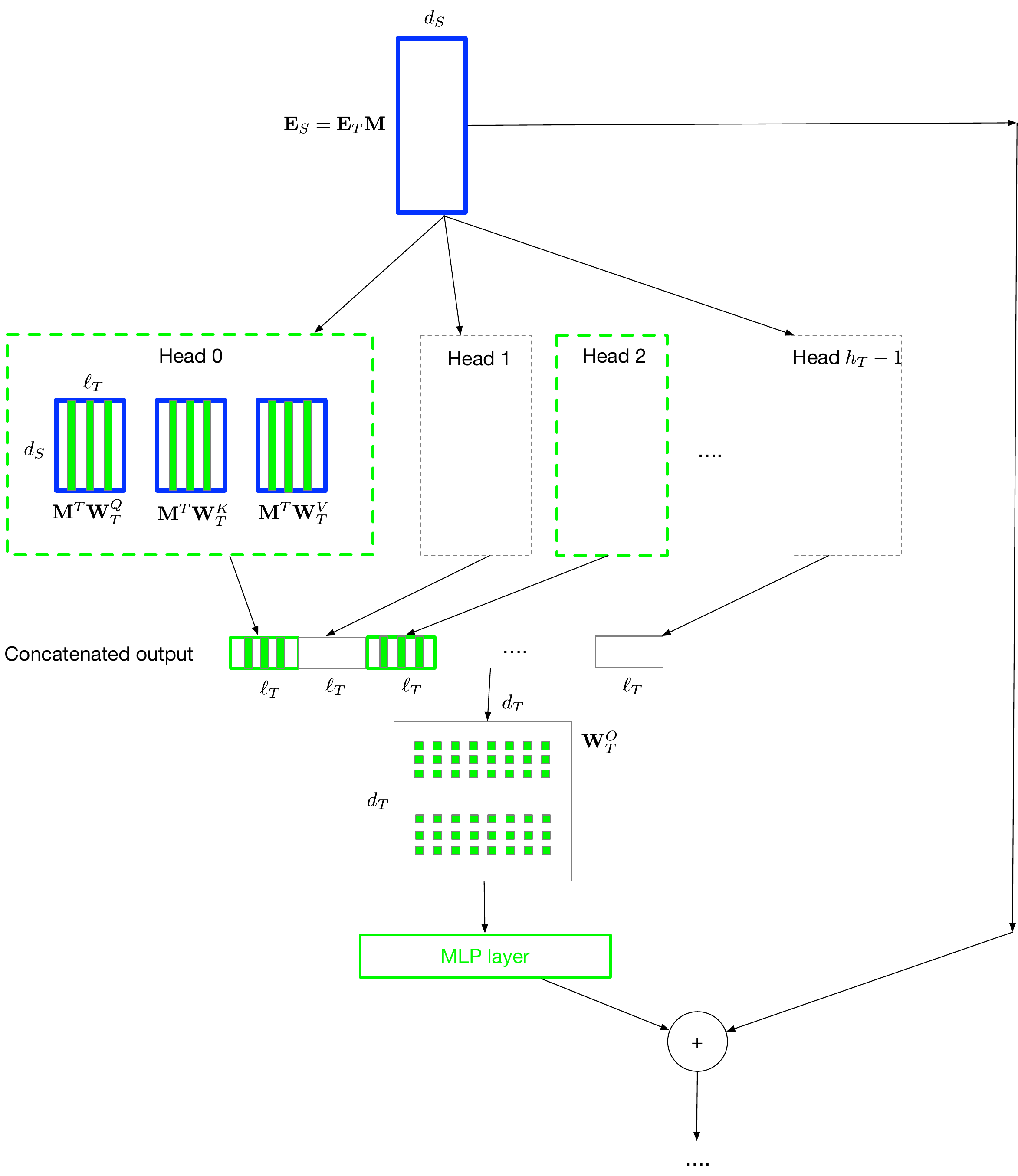}
\vspace{.1in}
\caption{
GUIDE uses $M^T$ to transform $Q, K, V$ matrices. Then Uniform Selection is used to take evenly-spaced green rows and columns to initialize the student’s first block.
}
\label{fig:algo1}
\end{figure}

Figure \ref{fig:algo1} illustrates how $M^T$ is used to transform $Q,K,V$ weight matrices, and evenly-spaced weights are selected from the remaining of the first transformer block by GUIDE.

\section{Experimental Results}
In this section, we experiment with \guide using traditional transformers of various sizes. In Section \ref{subsec:setup}, we describe the setting of our experiments. The main experimental results of will be presented in Section \ref{subsec:main_res}. In Section \ref{subsec:combining}, we show that GUIDE can be combined with standard distillation, giving improved performance. Finally, we present the ablation study and the results when initializing intermediate transformer blocks in Section \ref{subsec:middle_layers}.

\subsection{Experimental Setup}\label{subsec:setup}
In all of our experiments, we pre-train a decoder-only transformer on the public Colossal Cleaned Common Crawl (C4) dataset \cite{raffel2020c4} using the NanoDO training framework by \cite{nanodo}. The configurations of the models presented in this section are shown in Table \ref{tab:config}.

\begin{table}[H]
\setlength{\tabcolsep}{4pt}
\caption{Model configurations for the teacher and the student models in our experiments.}
\label{tab:config}

\begin{center}
\begin{small}
\begin{tabular}{lllllll}
\toprule
\bf{Type} & \bf{Model} & \bf{Model} & \bf{Num} & \bf{Num} & \bf{Head} & \bf{FFN} \\ 
 & \bf{Size} & \bf{Dims} & \bf{Layers} & \bf{Heads} & \bf{Dims}& \bf{Dims} \\
\midrule
Student         &400M  &960 &23 &30 &32 &5,760 \\
Student         &1B  &1,728 &25 &36 &48 &6,912 \\
Teacher         &4.2B  &3,072 &36 &48 &64 &12,288 \\
\bottomrule
\end{tabular}
\end{small}
\end{center}
\end{table}

All models have the same vocabulary size of 32,000, and a context length of 2048 tokens. It takes about $8.5$ hours, $14$ hours, and 3.5 days to train the 400M student, 1B student, and 4.2B teacher models respectively, on the Google TPU Trillium (v6e) chips. Table \ref{tab:training_config} shows the training configurations of the above models.

\begin{table}[H]
\caption{Training configurations for the student and teacher models.}
\label{tab:training_config}
\begin{center}
\begin{small}
\begin{tabular}{llllll}
\toprule
\bf{Type} & \bf{Model} & \bf{Batch} & \bf{Num TPU} & \bf{Training}\\ 
 & \bf{Size} & \bf{Size} & \bf{Chips} & \bf{Steps} \\
\midrule

Student         &400M  &192 &64 &50,000   \\
Student         &1B  &192 &64 &50,000   \\
Teacher         &4.2B  &1,024 &128 &50,000   \\
\bottomrule
\end{tabular}
\end{small}
\end{center}
\end{table}

\subsection{Main Results}\label{subsec:main_res}
We compare \guide with a student trained from scratch with random initialization, and variants of Uniform Selection by \cite{xu2023init}. Interestingly, using the ``first-$N$ selection'' option for layer selection as suggested by \cite{xu2023init} performs quite poorly, causing the student to perform worse than using random initialization. Taking a few evenly-spaced layers (1 or 2) works better and is used as the main baseline in our study. The final perplexities of the student model are reported in Tables \ref{tab:400m} and  \ref{tab:1b}.

\begin{table}[h]
\caption{Comparison between GUIDE, Uniform Select, and Random Initialization for the 400M parameter student model when using the 4.2B parameter teacher model.}
\label{tab:400m}
\begin{center}
\begin{small}
\begin{tabular}{lllll}
\toprule
\bf{Initialization} & \bf{Perplexity} & \textbf{Teacher-Student}\\ 
 \bf{Algorithm} & & \bf{Gap Reduction}\\
 &  & \bf{(\%)} \\
\midrule
Random Initialization   & 15.915$\pm$0.015&  N/A  \\
Uniform Select. (first-$N$)   & 15.967$\pm$0.015  &  -0.82  \\
Uniform Select. (1 layer)   & 14.458$\pm$0.013  &  23.15  \\
Uniform Select. (2 layers)   & 14.498$\pm$0.013  & 22.51   \\
\textbf{GUIDE}  &  \textbf{14.245$\pm$0.013} &  \textbf{26.53}  \\
\midrule
Teacher   & 9.621$\pm$0.004  &  N/A  \\
\bottomrule
\end{tabular}
\end{small}
\end{center}
\end{table}


\begin{table}[h]
\caption{Comparison between GUIDE, Uniform Select, and Random Initialization for the 1B parameter student model when using the 4.2B parameter teacher model.}
\label{tab:1b}
\begin{center}
\begin{small}
\begin{tabular}{lllll}
\toprule
\bf{Initialization} & \bf{Perplexity} & \textbf{Teacher-Student}\\ 
\bf{Algorithm} & & \bf{Gap Reduction}\\
 &  & \bf{(\%)} \\
\midrule
Random Initialization   & 13.382$\pm$0.012  &  N/A  \\
Uniform Select. (first-$N$)   & 14.088$\pm$0.013  &  -18.77  \\
Uniform Select. (1 layer)   & 12.459$\pm$0.011  &  24.54  \\
Uniform Select. (2 layers)   & 12.491$\pm$0.011  & 23.70   \\
\textbf{GUIDE}  &  \textbf{12.438$\pm$0.011} &  \textbf{25.11}  \\
\midrule
Teacher   & 9.621$\pm$0.004  &  N/A  \\
\bottomrule
\end{tabular}
\end{small}
\end{center}
\end{table}

\begin{figure}[H]
\centering 
\vspace{.1in}
\includegraphics[width=0.45\textwidth]{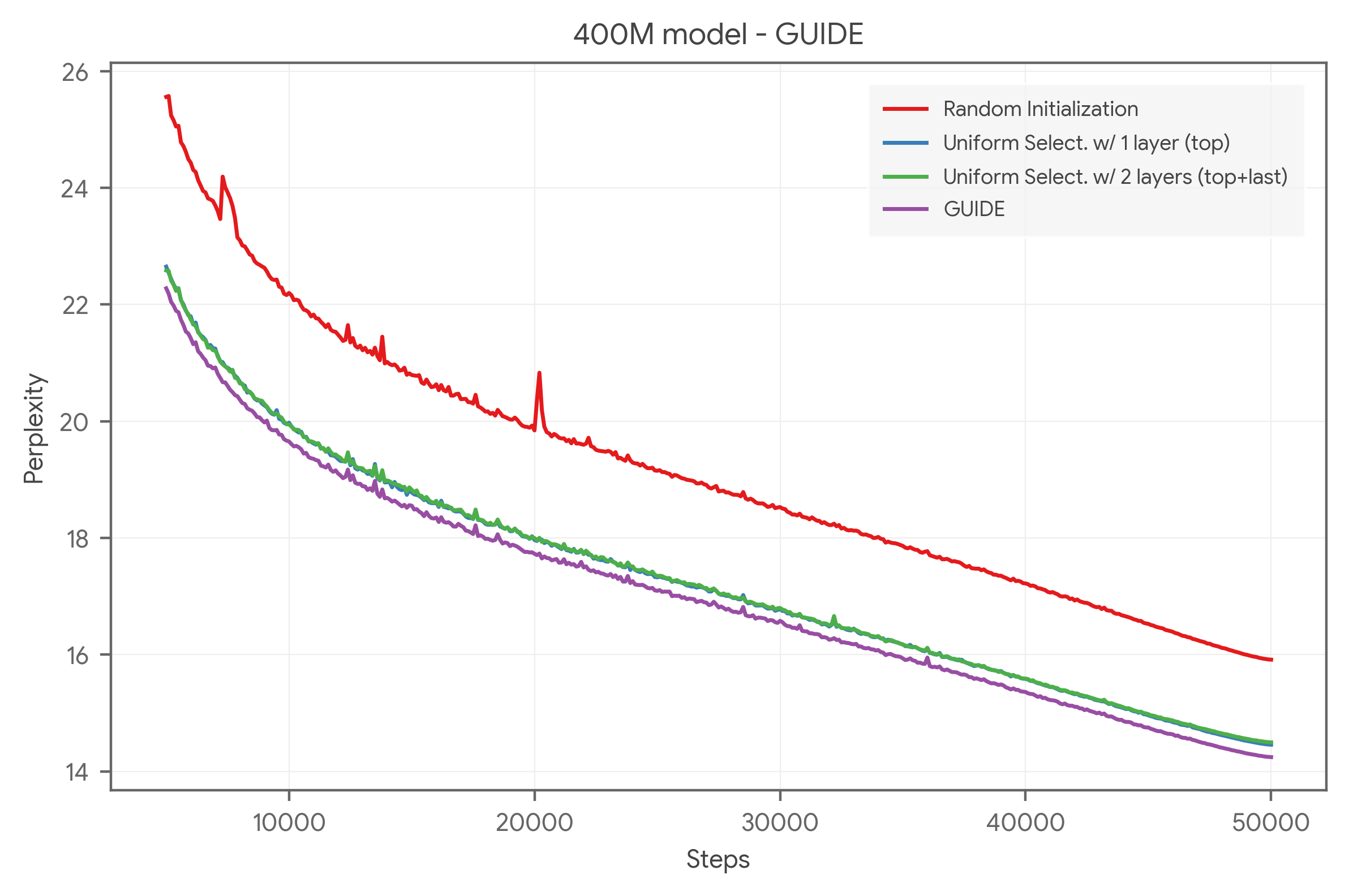}
\vspace{.1in}
\caption{Training trajectory of the 400M parameter student model with GUIDE and other initialization methods. x-axis is the step number, and y-axis is the perplexity on the evaluation dataset.}
\label{fig:400m_trajectory}
\end{figure}

\begin{figure}[h]
\centering 
\vspace{.1in}
\includegraphics[width=0.45\textwidth]{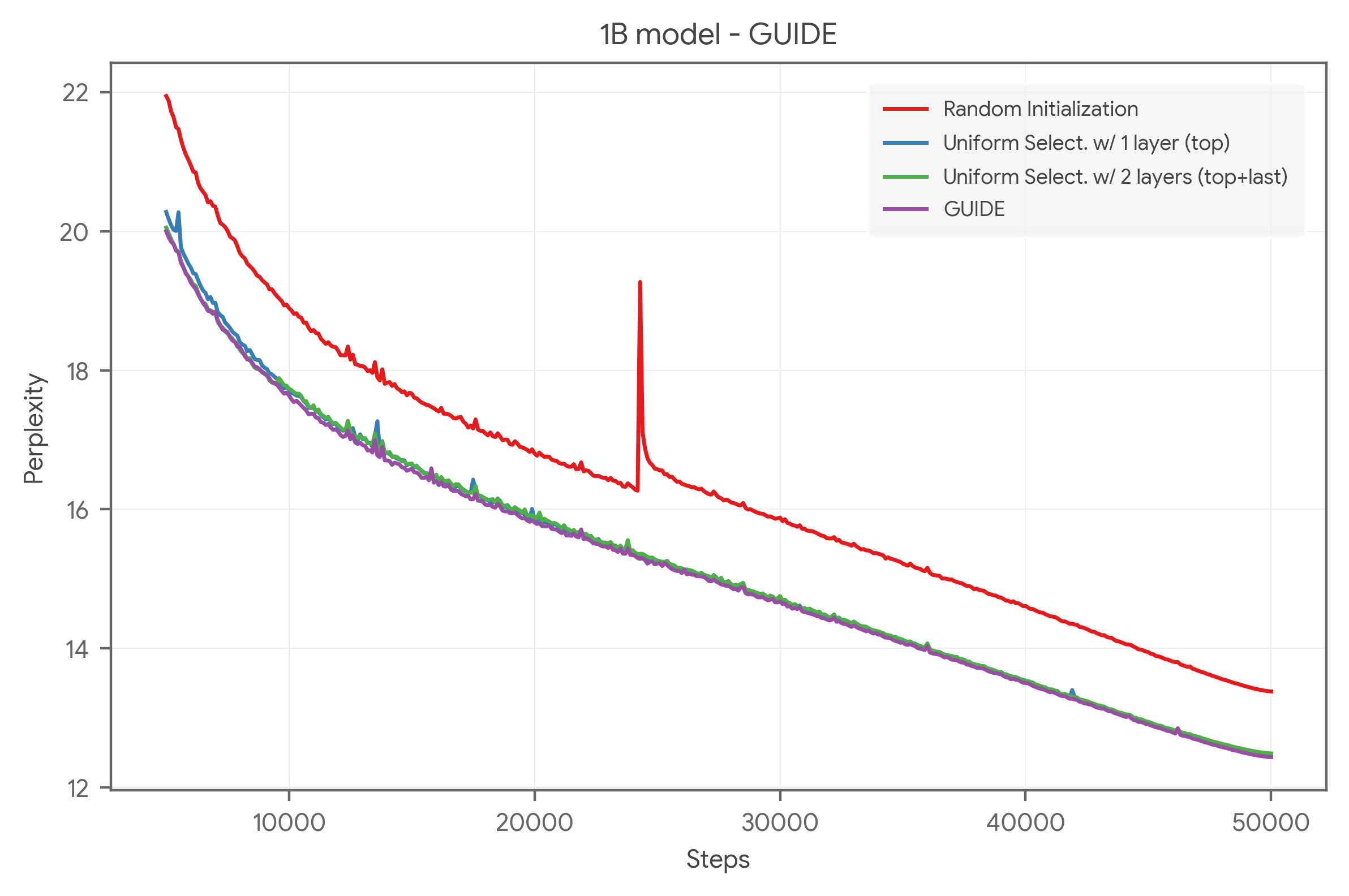}
\vspace{.1in}
\caption{Training trajectory of the 1B parameter student model with GUIDE and other initialization methods. x-axis is the step number, and y-axis is the perplexity on the evaluation dataset.}
\label{fig:1b_trajectory}
\end{figure}

We plot the perplexity of the student models on the training trajectory in Figures \ref{fig:400m_trajectory} and \ref{fig:1b_trajectory}. In both cases, GUIDE consistently outperforms existing approaches throughout the training process. In summary, GUIDE is able to reduce the teacher-student perplexity gap by $26.52\%$ and $25.11\%$ for the 400M and 1B models, respectively.

\begin{figure}[H]
\centering 
\vspace{.1in}
\includegraphics[width=0.45\textwidth]{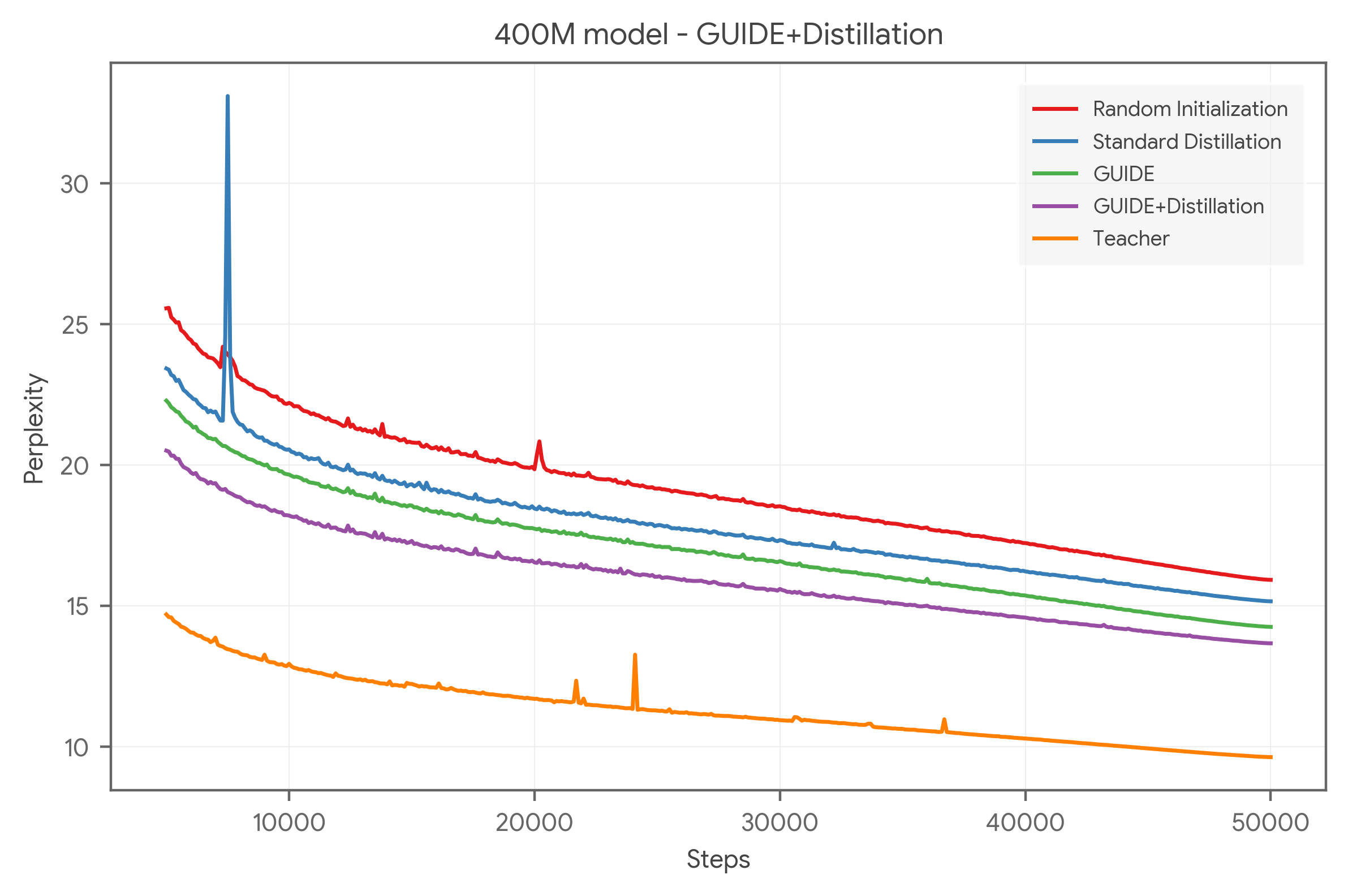}
\vspace{.1in}
\caption{Combining Knowledge Distillation with GUIDE when training 400M model. }
\label{fig:400m_dist}
\end{figure}

\begin{figure}[h]
\centering 
\vspace{.1in}
\includegraphics[width=0.45\textwidth]{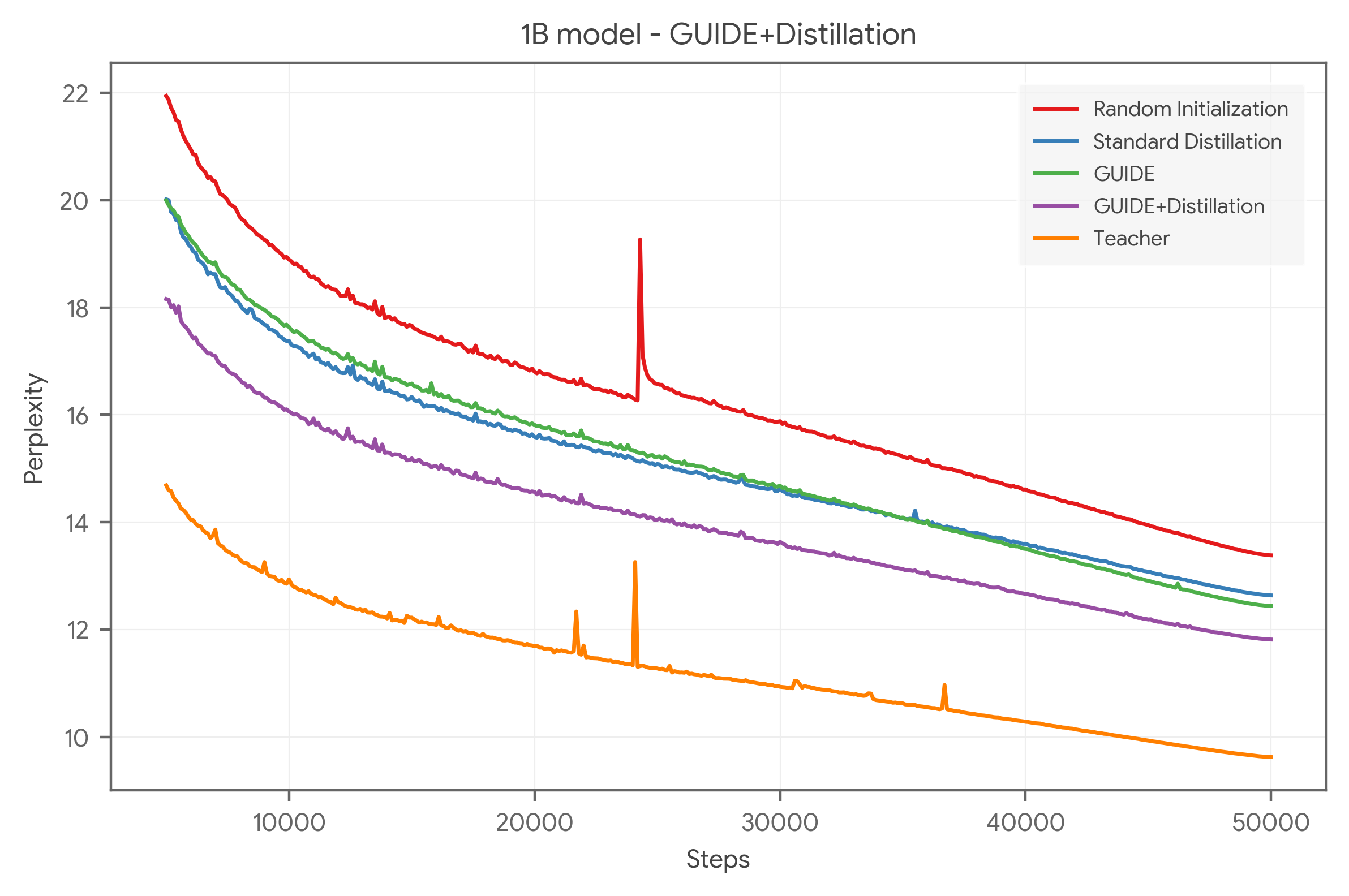}
\vspace{.1in}
\caption{Combining Knowledge Distillation with GUIDE when training 1B model.}
\label{fig:1b_dist}
\end{figure}

Observe that there are several ``loss spikes'' during the training trajectory of the non-\guide models in Figures \ref{fig:400m_trajectory}, \ref{fig:1b_trajectory}, \ref{fig:400m_dist}, and \ref{fig:1b_dist}. In fact, this phenomenon has been reported to negatively affect the training of large models \cite{palm}. Notably, GUIDE does not suffer from any big loss spike, demonstrating the stability of our approach.

\subsection{Combining with Knowledge Distillation}\label{subsec:combining}

Here we show that applying GUIDE before using Knowledge Distillation (KD) will help boost the quality of the student significantly. In our distillation experiments, we use the same 4B parameter teacher model for both, \guide and knowledge distillation. For KD, we load the teacher into memory for inference and freeze its parameters. The teacher is then used to generate its per-token distributions that the student model is forced to match. More specifically, we minimize the following loss:
$$\mathcal{L}_{\text{total}} = \mathcal{L}_{\text{pred}} + \alpha   \mathcal{L}_{\text{distill}},  $$
where $\mathcal{L}_{\text{pred}}$ is the usual next-token prediction loss and $\mathcal{L}_{\text{distill}}$ is the distillation loss, which is defined as the cross entropy between the per-token distributions returned by the student and the teacher. We set the distillation weight $\alpha := 0.5$ in our experiments.

The performance of the students when using GUIDE + KD is reported in Tables \ref{tab:400m_dist} and \ref{tab:1b_dist}. GUIDE itself is capable of transferring more information from the teacher to the student than KD alone. Moreover, the gap reduction of the combined GUIDE + KD setup is almost equal to the sum gap reductions of using the two methods separately. This suggests that GUIDE and KD work synergistically, with each bringing distinct improvements.

\begin{table}[h]
\caption{Combining GUIDE and Knowledge Distillation for the 400M parameter student model.}
\label{tab:400m_dist}
\begin{center}
\begin{small}
\begin{tabular}{lllll}
\toprule
\bf{Model} & \bf{Perplexity} & \textbf{Teacher-Student}\\ 
 & & \bf{Gap Reduction} \\
 & & \bf{(\%)}\\
\midrule
KD   & 15.154$\pm$0.014 & 12.10  \\		
GUIDE   & 14.246$\pm$0.013  & 26.53   \\ 
GUIDE + KD   &  13.662$\pm$0.012 &  35.80  \\		
\bottomrule
\end{tabular}
\end{small}
\end{center}
\end{table}

\begin{table}[h]
\caption{Combining GUIDE and Knowledge Distillation for the 1B parameter student model.}
\label{tab:1b_dist}
\begin{center}
\begin{small}
\begin{tabular}{lllll}
\toprule
\bf{Model} & \bf{Perplexity} & \textbf{Teacher-Student } \\ 
 & & \bf{Gap Reduction} \\
 & & \bf{(\%)}\\
\midrule
KD   & 12.636$\pm$0.011 & 19.86  \\		
GUIDE   & 12.438$\pm$0.011  & 25.11   \\ 
GUIDE + KD   &  11.813$\pm$0.010 &  41.73  \\		
\bottomrule
\end{tabular}
\end{small}
\end{center}
\end{table}

\subsection{Ablation Study and Initializing Intermediate Layers}\label{subsec:middle_layers}
Our most interesting observation here is that initializing the first layer is crucial. This will significantly improve the student's performance compared to initializing the student's embedding table alone. However, initializing more intermediate layers after this point would not help much in general. Furthermore, our results show that the ``first-$N$'' strategy suggested by \cite{xu2023init} for layer selection performs poorly on standard transformers. The complete results for the 400M and 1B models are presented in Tables \ref{tab:400m_layers} and \ref{tab:1b_layers}.

\begin{table}[h]
\caption{Performance of GUIDE with different strategies for layer selection on the 400M parameter student model. We find that applying GUIDE on the embedding table and the top most layer gives the best results.}
\label{tab:400m_layers}
\begin{center}
\begin{small}
\begin{tabular}{lllll}
\toprule
\bf{GUIDE} & \bf{Perplexity} & \textbf{Teacher-Student}\\ 
 & & \bf{Gap Reduction}\\
 & & \bf{(\%)}\\
\midrule
Embed. Table only  & 14.453$\pm$0.013  &  23.24  \\
\\
Embed. Table + \\
\hspace{6pt}\textbf{1 layer (top)}  & \textbf{14.246$\pm$0.013}  & \textbf{26.53 }  \\ 
\hspace{6pt}2 layers (top+last)   & 14.317$\pm$0.013  & 25.40  \\	
\hspace{6pt}3 layers    & 14.268$\pm$0.013 & 26.18   \\	
\hspace{6pt}4 layers   & 14.356$\pm$0.013   &  24.79 \\	
\hspace{6pt}8 layers    & 14.509$\pm$0.013 & 22.35 \\	
\hspace{6pt}first-$N$ layers    &  15.501$\pm$0.014 & 6.58 \\	
\bottomrule
\end{tabular}
\end{small}
\end{center}
\end{table}

\begin{table}[h]
\caption{Performance of GUIDE with different strategies for layer selection on the 1B parameter student model. Consistent with Table \ref{tab:400m_layers} we find that applying GUIDE on the embedding table and the top most layer gives the best results.}
\label{tab:1b_layers}
\begin{center}
\begin{small}
\begin{tabular}{lllll}
\toprule
\bf{GUIDE} & \bf{Perplexity} & \textbf{Teacher-Student  } \\ 
 & & \bf{Gap Reduction}\\
 & & \bf{(\%)}\\
\midrule
Embed. Table only  & 12.866$\pm$0.012  &  13.73  \\	
\\
Embed. Table + \\
\hspace{6pt}\textbf{1 layer (top)}  &  \textbf{12.438$\pm$0.011} &   \textbf{25.12} \\ 
\hspace{6pt}2 layers (top+last)   & 12.612$\pm$0.011  &  20.50  \\	
\hspace{6pt}3 layers    & 12.505$\pm$0.011  &  23.34  \\	
\hspace{6pt}4 layers    &  12.609$\pm$0.011  & 20.58  \\	
\hspace{6pt}8 layers & 12.630$\pm$0.011 & 20.01 \\	
\hspace{6pt}first-$N$ layers    &  14.037$\pm$0.013& -17.40 \\	
\bottomrule
\end{tabular}
\end{small}
\end{center}
\end{table}

\section{Discussion}
In this section, we will summarize our results and observations from the previous section. We started with describing the model configuration and the training setup. Our choice of student and teacher model sizes is two orders of magnitude larger than the closest related work of \cite{xu2023init}. The student models were trained for $\approx$ 20B tokens and the teacher model was trained with $\approx$ 100B tokens.

As mentioned in Section \ref{subsec:main_res}, the baseline approach of `Uniform Selection' and matching the `first-$N$' layers of the teacher model with the student model as described \cite{xu2023init} performs poorly (where $N$ is the number of student layers). We suspect that since the authors report the results of their method on student and teacher models of sizes 5M and 22M parameters, it is possible that their initialization scheme does not transfer well to the much larger models such as LLMs of today. In fact, we find that choosing a smaller number of layers to match by evenly spacing them in the teacher model works better, and we use that as an `enhanced' baseline in our experiments.

In Tables \ref{tab:400m} and \ref{tab:1b} we see that \guide gives a 26.53\% and 25.11\% reduction teacher-student quality gap for the 400M and 1B param baselines, respectively. It significantly outperforms our `enhanced' baselines of Uniform Select with 1 and 2 layers.

In Section \ref{subsec:combining}, we explored combining \guide with standard knowledge distillation. Interestingly, \guide leads to additive gains on top of knowledge distillation, while also getting better model quality by itself than when knowledge distillation. The unique gains provide an interesting opportunity for those training models from scratch to leverage the teacher's knowledge at initialization time as well using \guide, apart from learning from it during training with standard KD.

Finally, we note that we get the best results when initializing the embedding table and the first / top layer from the teacher, as seen in Tables \ref{tab:400m_layers} and \ref{tab:1b_layers}. Matching more than 1 layer seems to lead to degradation in performance (except when matching 3 layers), suggesting that there is room for improvement in the method.

To summarize, given that \guide has no training and inference overhead, when a larger pre-trained model is present, applying \guide on a student model's embedding table and first layer is likely to improve model performance substantially. Combining \guide with knowledge distillation would help the student learn from the teacher both in the parametric space and the output space.  

\section{Conclusion}
In this paper, we introduce \guide, which is a novel algorithmic efficiency technique that helps improve model quality without any impact to the model size or latency.

We ran a host of experiments to demonstrate the efficacy of our method, and show that not only does \guide beat the baselines from existing literature, it also beats our `enhanced' versions of those baselines. \guide also combines very well with vanilla knowledge distillation method where we see that the quality gains from \guide are nearly additive on top of those from distillation. This strong result implies that \guide brings model quality gains that are distinct from the standard knowledge distillation approach. Furthermore, applying \guide alone led to better model quality than applying knowledge distillation alone. 

Therefore, when a large pre-trained models is available, using \guide to initialize smaller models before training is likely to provide virtually free improvements to model quality. In terms of future work, we aim to further improve \guide by developing novel variants that help us extract even more information from the teacher models during initialization and training.
\clearpage

\bibliographystyle{apalike}
\bibliography{guide}

\begin{thebibliography}{}

\bibitem[Chen et~al., 2022]{chen-etal-2022-bert2bert}
Chen, C., Yin, Y., Shang, L., Jiang, X., Qin, Y., Wang, F., Wang, Z., Chen, X., Liu, Z., and Liu, Q. (2022).
\newblock bert2{BERT}: Towards reusable pretrained language models.
\newblock In Muresan, S., Nakov, P., and Villavicencio, A., editors, {\em Proceedings of the 60th Annual Meeting of the Association for Computational Linguistics (Volume 1: Long Papers)}, pages 2134--2148, Dublin, Ireland. Association for Computational Linguistics.

\bibitem[Chowdhery et~al., 2023]{palm}
Chowdhery, A., Narang, S., Devlin, J., Bosma, M., Mishra, G., Roberts, A., Barham, P., Chung, H.~W., Sutton, C., Gehrmann, S., Schuh, P., Shi, K., Tsvyashchenko, S., Maynez, J., Rao, A., Barnes, P., Tay, Y., Shazeer, N., Prabhakaran, V., Reif, E., Du, N., Hutchinson, B., Pope, R., Bradbury, J., Austin, J., Isard, M., Gur-Ari, G., Yin, P., Duke, T., Levskaya, A., Ghemawat, S., Dev, S., Michalewski, H., Garcia, X., Misra, V., Robinson, K., Fedus, L., Zhou, D., Ippolito, D., Luan, D., Lim, H., Zoph, B., Spiridonov, A., Sepassi, R., Dohan, D., Agrawal, S., Omernick, M., Dai, A.~M., Pillai, T.~S., Pellat, M., Lewkowycz, A., Moreira, E., Child, R., Polozov, O., Lee, K., Zhou, Z., Wang, X., Saeta, B., Diaz, M., Firat, O., Catasta, M., Wei, J., Meier-Hellstern, K., Eck, D., Dean, J., Petrov, S., and Fiedel, N. (2023).
\newblock Palm: scaling language modeling with pathways.
\newblock {\em J. Mach. Learn. Res.}, 24(1).

\bibitem[Dehghani et~al., 2022]{dehghani2022efficiency}
Dehghani, M., Tay, Y., Arnab, A., Beyer, L., and Vaswani, A. (2022).
\newblock The efficiency misnomer.
\newblock In {\em ICLR}.

\bibitem[Devvrit et~al., 2024]{devvrit2024matformer}
Devvrit, Kudugunta, S., Kusupati, A., Dettmers, T., Chen, K., Dhillon, I., Tsvetkov, Y., Hajishirzi, H., Kakade, S., Farhadi, A., and Jain, P. (2024).
\newblock Matformer: Nested transformer for elastic inference.

\bibitem[Frankle and Carbin, 2019]{frankle2019lottery}
Frankle, J. and Carbin, M. (2019).
\newblock The lottery ticket hypothesis: Finding sparse, trainable neural networks.

\bibitem[Gale et~al., 2019]{gale2019sparsity}
Gale, T., Elsen, E., and Hooker, S. (2019).
\newblock The state of sparsity in deep neural networks.
\newblock {\em arXiv}, 1902.09574.

\bibitem[Hinton et~al., 2014]{hinton2015distillation}
Hinton, G., Vinyals, O., and Dean, J. (2014).
\newblock Distilling the knowledge in a neural network.
\newblock In {\em Deep Learning Workshop (NeurIPS)}.

\bibitem[Jacob et~al., 2018]{jacob2018quantization}
Jacob, B., Kligys, S., Chen, B., Zhu, M., Tang, M., Howard, A., Adam, H., and Kalenichenko, D. (2018).
\newblock Quantization and training of neural networks for efficient integer-arithmetic-only inference.
\newblock In {\em CVPR}, pages 2704--2713.

\bibitem[Jiao et~al., 2020]{tinybert}
Jiao, X., Yin, Y., Shang, L., Jiang, X., Chen, X., Li, L., Wang, F., and Liu, Q. (2020).
\newblock {T}iny{BERT}: Distilling {BERT} for natural language understanding.
\newblock In Cohn, T., He, Y., and Liu, Y., editors, {\em Findings of the Association for Computational Linguistics: EMNLP 2020}, pages 4163--4174, Online. Association for Computational Linguistics.

\bibitem[Kim et~al., 2023]{kim2023embeddistill}
Kim, S., Rawat, A.~S., Zaheer, M., Jayasumana, S., Sadhanala, V., Jitkrittum, W., Menon, A.~K., Fergus, R., and Kumar, S. (2023).
\newblock Embeddistill: A geometric knowledge distillation for information retrieval.

\bibitem[Krishnamoorthi, 2018]{krishnamoorthi2018quantization}
Krishnamoorthi, R. (2018).
\newblock {Quantizing deep convolutional networks for efficient inference: A whitepaper}.
\newblock {\em arXiv}, 1806.08342.

\bibitem[Lin et~al., 2021]{weight_dist}
Lin, Y., Li, Y., Wang, Z., Li, B., Du, Q., Xiao, T., and Zhu, J. (2021).
\newblock distillation: Transferring the knowledge in neural network parameters.
\newblock In Zong, C., Xia, F., Li, W., and Navigli, R., editors, {\em Proceedings of the 59th Annual Meeting of the Association for Computational Linguistics and the 11th International Joint Conference on Natural Language Processing (Volume 1: Long Papers)}, pages 2076--2088, Online. Association for Computational Linguistics.

\bibitem[Liu et~al., 2024]{nanodo}
Liu, P.~J., Novak, R., Lee, J., Wortsman, M., Xiao, L., Everett, K., Alemi, A.~A., Kurzeja, M., Marcenac, P., Gur, I., Kornblith, S., Xu, K., Elsayed, G., Fischer, I., Pennington, J., Adlam, B., and Dickstein, J.-S. (2024).
\newblock Nanodo: A minimal transformer decoder-only language model implementation in {JAX}.

\bibitem[Menghani, 2023]{menghani2023efficientdl}
Menghani, G. (2023).
\newblock Efficient deep learning: A survey on making deep learning models smaller, faster, and better.
\newblock {\em ACM Computing Surveys}.

\bibitem[Menghani et~al., 2025]{menghani2025laurel}
Menghani, G., Kumar, R., and Kumar, S. (2025).
\newblock Laurel: Learned augmented residual layer.

\bibitem[Raffel et~al., 2020]{raffel2020c4}
Raffel, C., Shazeer, N., Roberts, A., Lee, K., Narang, S., Matena, M., Zhou, Y., Li, W., and Liu, P.~J. (2020).
\newblock Exploring the limits of transfer learning with a unified text-to-text transformer.
\newblock {\em J. Mach. Learn. Res.}, 21(1).

\bibitem[Romero et~al., 2014]{romero2014fitnets}
Romero, A., Ballas, N., Kahou, S.~E., Chassang, A., Gatta, C., and Bengio, Y. (2014).
\newblock Fitnets: Hints for thin deep nets. arxiv 2014.
\newblock {\em arXiv preprint arXiv:1412.6550}.

\bibitem[Samragh et~al., 2024]{Samragh2024ScalingSA}
Samragh, M., Mirzadeh, I., Vahid, K.~A., Faghri, F., Cho, M., Nabi, M., Naik, D., and Farajtabar, M. (2024).
\newblock Scaling smart: Accelerating large language model pre-training with small model initialization.
\newblock In {\em ENLSP}.

\bibitem[Sanh et~al., 2019]{sanh2019distilbert}
Sanh, V., Debut, L., Chaumond, J., and Wolf, T. (2019).
\newblock Distilbert, a distilled version of bert: smaller, faster, cheaper and lighter.

\bibitem[Shleifer and Rush, 2020]{sum_dist}
Shleifer, S. and Rush, A.~M. (2020).
\newblock Pre-trained summarization distillation.
\newblock {\em CoRR}, abs/2010.13002.

\bibitem[Sze et~al., 2017]{sze2017survey}
Sze, V., Chen, Y., Yang, T., and Emer, J.~S. (2017).
\newblock Efficient processing of deep neural networks: {A} tutorial and survey.
\newblock {\em Proc. {IEEE}}, 105(12):2295--2329.

\bibitem[Vaswani et~al., 2017]{transformer}
Vaswani, A., Shazeer, N., Parmar, N., Uszkoreit, J., Jones, L., Gomez, A.~N., Kaiser, L.~u., and Polosukhin, I. (2017).
\newblock Attention is all you need.
\newblock In Guyon, I., Luxburg, U.~V., Bengio, S., Wallach, H., Fergus, R., Vishwanathan, S., and Garnett, R., editors, {\em Advances in Neural Information Processing Systems}, volume~30. Curran Associates, Inc.

\bibitem[Wortsman et~al., 2022]{wortsman2022soup}
Wortsman, M., Ilharco, G., Gadre, S.~Y., Roelofs, R., Gontijo-Lopes, R., Morcos, A.~S., Namkoong, H., Farhadi, A., Carmon, Y., Kornblith, S., and Schmidt, L. (2022).
\newblock Model soups: averaging weights of multiple fine-tuned models improves accuracy without increasing inference time.

\bibitem[Xia et~al., 2024]{xia2024sheared}
Xia, M., Gao, T., Zeng, Z., and Chen, D. (2024).
\newblock Sheared {LL}a{MA}: Accelerating language model pre-training via structured pruning.
\newblock In {\em The Twelfth International Conference on Learning Representations}.

\bibitem[Xu et~al., 2023]{xu2023init}
Xu, Z., Chen, Y., Vishniakov, K., Yin, Y., Shen, Z., Darrell, T., Liu, L., and Liu, Z. (2023).
\newblock Initializing models with larger ones.

\bibitem[Zhong et~al., 2024]{zhong2024seeking}
Zhong, M., An, C., Chen, W., Han, J., and He, P. (2024).
\newblock Seeking neural nuggets: Knowledge transfer in large language models from a parametric perspective.
\newblock In {\em The Twelfth International Conference on Learning Representations}.

\end{thebibliography}

\end{document}